\documentclass[10pt, a4paper]{IEEEtran}

\usepackage[utf8]{inputenc}
\usepackage{anyfontsize} 

\usepackage[table,xcdraw]{xcolor}
\usepackage{amsmath}
\usepackage{amsfonts}
\usepackage{amssymb}
\usepackage{dsfont}

\let\emptyset\varnothing
\usepackage{graphicx}
\usepackage{balance}
\usepackage{afterpage} 
\usepackage{algorithm}
\usepackage{outlines}
\usepackage{cite}
\usepackage[caption=false]{subfig}

\usepackage{tabularx}

\usepackage{paul}
\usepackage{hyperref}
\usepackage{cleveref}
\crefname{problem}{problem}{problems}
\creflabelformat{problem}{(#2#1#3)}

\graphicspath{{./figures/}}

\newcommand{\emphb}[1]{\textbf{\emph{#1}}}

\newcommand{\nlw}[1][1]{\\[0.5ex]} 

\newcommand{\musec}{$\mu$s}

\mc{\Arc}{\mathcal{A}} 
\mc{\dom}{\succ}
\mc{\wdom}{\succeq}
\mc{\Nfe}{N_{f}}
\mc{\Nruns}{N_{r}}
\mc{\popsize}{\lambda}
\mc{\Iea}{\mathcal{I}_{\epsilon,+}}
\mc{\Iem}{\mathcal{I}_{\epsilon,\times}}
\mc{\NDS}{\text{NDSort}}
\mc{\bmin}{\underline{m}}
\mc{\bmax}{\overline{m}}
\mc{\tf}{\tilde{f}}
\newcommand{\epsind}{$\epsilon$-indicator}
\newcommand{\tdea}{$\theta$-DEA}

\author{Paul Dufoss\'e\inst{1, 2} \and Cyrille Enderli\inst{2}}

\begin{document}

\title{Finding optimal Pulse Repetition Intervals with Many-objective Evolutionary Algorithms}
\author{\IEEEauthorblockN{Paul Dufoss\'e\IEEEauthorrefmark{1}\IEEEauthorrefmark{2},
		Cyrille Enderli\IEEEauthorrefmark{1}\\
		\IEEEauthorblockA{\IEEEauthorrefmark{1}Thales Defense Mission Systems, France}\\
		\IEEEauthorblockA{\IEEEauthorrefmark{2}Randopt team, Inria Paris-Saclay and CMAP, \'Ecole Polytechnique, France}
}}
\maketitle
	
\begin{abstract}
We consider the problem of finding Pulse Repetition Intervals allowing the best compromises mitigating range and velocity ambiguities in a Pulse-Doppler radar system.
This problem has been previously proposed as a Many-objective optimization problem. We run several Evolutionary Algorithms to obtain an exhaustive set of candidate solutions.
We study algorithm performance relative to common metrics in Many-objective optimization.
The aggregated dataset of Pareto optimal points reveals great improvement over the existing one.
\new{This approach is appealing for the radar waveform designer as it encompass already found solutions from the literature and extend them with a much more diverse set.}
\end{abstract}

\done{merge Intro and Sec 1, remove preliminaries from sec 2}
\done{use cref}
\done{cut references for page limit}
\done{add grid in boxplot figure}

\section{Introduction}

Airborne Pulse-Doppler radars are complex systems and have been further developed for decades by engineers. The radar waveform (WF) is based on diverse Pulse Repetition Intervals (PRIs) or equivalent Pulse Repetion Frequencies (PRFs).
Almost all Pulse-Doppler radars operate at medium PRF (between 3 kHz to 30 kHz) hence produce ambiguous range and Doppler measures.
The common technique to overcome this situation is to construct a WF consisting of \new{several train of pulses} with different PRIs.
Researchers already found several solutions with the help of various search heuristics.

In \cite{simpson1988prf} \del{the author}\new{J. Simpson} finds PRFs guided by a constraint programming approach: first generating a set of feasible candidates, then sequentially testing the candidates with 3 scoring passes, finally examining blind zones plots.
The AN/APG-69 radar system \new{with constant pulse width} is chosen as a baseline and the paper ends mentioning a single choice solution of 8 PRFs.

In 2003 \cite{alabaster2003medium} C. Alabaster and E. Hughes apply an Evolutionary Algorithm (EA) to identify near-optimal vectors of 8 or 9  PRFs for  a practical fire control radar system. E. Hughes would later propose a similar problem under the framework of Many-objective optimization that we address in this paper.

In \cite{ahn2011medium} \new{Ahn et al.}\del{the authors} propose a model that is similar to the one considered here, with a constant duty cycle but a somehow different parameters setting (smaller dwell time, smaller Doppler interval \dots) and aims to minimize the blind zones while ensuring full decodability in the zone of interest. A single solution with 8 PRIs is found using a Simulated Annealing approach.

\done{paragraphe introductif MO ici}
We propose to revisit a radar PRI design problem proposed 13 years ago to the Evolutionary Multi-objective Optimization (EMO) community and stated as a Many-objective Optimization problem\footnote{in the EMO literature the term Multi-objective is replaced with Many-objective when the number of objectives is $\geq 4$} (MaOP). 
Multi-objective Evolutionary Algorithms (MOEAs) are meta-heuristic optimization algorithms specifically designed to handle such problems. They \new{can approximate irregular, nonsmooth Pareto Fronts.}\del{solve non-convex, multi-modal, and non-differentiable problems.}
We generate candidate solutions with a bench of \new{contemporary} evolutionary algorithms and assess their performance on our application. Our contributions can be summarized as follows:
\begin{outline}
	\1 Improving the solution set for the PRI problem at hand\del{ (we found new optimal points)}
	\1 Pareto Front Analysis that allows the radar designer to better understand the trade-offs
	\1 Benchmarking EMO algorithms on a real-world problem
\end{outline}

\done{review this with new section formatting}
The paper is organised as follows: in \Cref{sec_problem} we remind the considered problem and the EMO framework, then we detail the algorithms and metrics used in \Cref{sec_perf}. Experimental results and analysis are presented in \Cref{sec_exp} and \Cref{sec_future} concludes the work and possible extensions.

\subsection*{Notations}
We denote by $\x$ an arbitrary real column-vector and $x_i$ its $i$-th coordinate. It is also called a \emph{point} in the $D$-dimensional search space and $\Points$ is a set of points $\{\x_1, \dots, \x_{\#\Points}\}$ and $\#\Points$ its cardinal. We denote as $f(\Points) = \{f(\x); \; \x \in \Points \}$ the Pareto Front induced by the set $\Points$. The notion of set used here may not be confused with the common meaning of a set of PRIs, represented here as a the $D$-dimensional vector.

\section{Problem definition}\label{sec_problem}

In 2007 Evan J. Hughes introduced a \del{real engineering}\new{WF design} problem to the EMO community \cite{radar2007hughes}. We recall the problem goals and key parameters, and how it fits the Many-objective optimization framework.

\subsection{Finding optimal Pulse Repetition Intervals}\label{subsec_problem}

\done{améliorer avec Cyrille (aussi remarque Reviewer 1)}
\todo{add nb of pulses in burst, space charging, maybe in the table}

 The goal is to mitigate the range and Doppler ambiguities. It also accounts for blind zones and total transmission time (or dwell time). 
The problem is to find a vector \new{of 4 to 12 PRIs between 50 and 150 \musec\ and consider a time quantization of 0.1 \musec.}
The simulation\done{pas clair} accepts any continuous value \del{in the range $[500,\,1500]$}\done{pas mentionner} as input and then does implicit rounding.
The performance measure is casted as a multivariate, multivalued real \del{continuous}\done{not here}function $f: \, \R^D \rightarrow \R^M$.  It has $M = 9$ objectives and varying dimension for the decision variables $D \in [4, 12]$.
We recall below what are the different objectives of the MaOP; an exhaustive list of the radar model's characteristics is reproduced in \Cref{radar:parameters} for the reader's convenience and more details about the objective function can be found in \cite{radar2007hughes}.

The process of finding the true, non-ambiguous range and velocity of the target, also called decodability \cite{kinghorn1997decodability}\done{reference ici}, takes into account possible measurements errors on the corresponding ambiguous measurements. \new{Such errors can lead to \emph{ghost targets} that are reflection of an actual target and should be properly rejected by the radar signal processing algorithm.}
\done{explaining ghosting (asked by reviewer)}
The notion of blindness accounts for eclipsing, mainbeam clutter rejection and sidelobe clutter power. 
Hence the MaOP aims at:
\begin{outline}
	\1 Maximizing the admissible error in range allowing decodability without ghosting ($f_1, \, f_5$)
	\1 Maximizing the admissible error in velocity allowing decodability without ghosting ($f_2, \, f_6$)
	\1 Maximizing the size of the clutter patch in range that can be tolerated before blind ranges occurs ($f_3, \, f_7$)
	\1 Maximizing the size of the clutter patch in velocity that can be tolerated before blind velocities occurs ($f_4, \, f_8$)
\end{outline}
Note that each underlying goal translates to 2 objectives as we optimize respectively for the median ($f_1$ to $f_4$) and minimum ($f_5$ to $f_8$) values because each range and velocity cell has an associated maximum target extent. This suggests that the objective functions are not totally independent.
Finally, it is desirable to spend as little time as possible in a given direction for discretion purposes. The last objective of the MaOP is to minimize the  dwell time ($f_9$).

\begin{table}
	\centering
	\begin{tabular}{l l}
		\textbf{Parameter} & \textbf{Value} \\
		\hline
		Carrier frequency & 9.97 GHz for first PRF, each following -30MHz \\
		Minimum PRI & 50 \musec \\		
		Maximum PRI & 150 \musec \\
		Compressed pulsewidth & 0.5 \musec \\
		Receiver recovery time & 1.0 \musec \\
		Range resolution & 75 m \\
		FFT size & 64 bins \\
		Duty cycle & 10\% fixed \\
		Maximum target dwell time & 50 ms \\
		Maximum target velocity & $ \pm 1500 \text{ ms}^{-1}$ \\
		Maximum detection range & 185.2 km \\
		Number of PRFs & 4 to 12 (10 in this paper) \\
		Number of PRFs for coincidence & 3 \\
		\hline
	\end{tabular}
	\caption{Characteristics of the \new{radar} model}
	\label{radar:parameters}
\end{table}

\subsection{Evolutionary Many Objective Optimization}\label{sec:emo}

From the description in \Cref{subsec_problem} we aim to solve the following MaOP :
\begin{equation}\label[problem]{problem_pri}
\text{(jointly) }\mini{\x \in \X} \, f(\x) = (-f_1(\x), \, \dots,\, -f_8(\x),\, f_9(\x))
\end{equation}
where $\X  = [500,\,1500]^D\del{ \subset \R^D}$ is the feasible space for the PRIs and we negate the first 8 objectives to fit a minimization problem.
\del{Here the terminology of Many Objective Optimization Problem (MaOP) is used because $M \geq 4$.}
The problem is said to be black-box as no other additional information (e.g. gradient\del{, hessian}) is available to the optimization algorithm.
MOEAS are \del{also} appealing as they output a set of Pareto-optimal solutions.
Popular MOEAs rely on non-dominated sorting to select preferred candidate solutions.
\subsubsection*{Pareto dominance}
Let $\x, \, \y \in \X$, then we say that $\x$ dominates $\y$, denoted $\x \dom \y$ if and only if $\forall i \, f_i(\x) \leq f_i(\y)$ and $\exists i \, f_i(\x) < f_i(\y)$. In a set $\Points$, a point $\x$ is said to be non-dominated if no point in $\Points$ dominates $\x$.

\del{It is well known that, w}\new{W}hen dealing with many objectives, the Pareto dominance relationship is not enough to ensure selection pressure, as most sampled points become non-dominated.
\new{We propose to look at several metrics to assess the performance of an optimizer.}

\section{Performance assessment}\label{sec_perf}
\done{title is OK}

It is not straightforward to evaluate an algorithm performance on MaOPs and even more difficult when the true Pareto Front is not known. \new{Several metrics have been proposed and lead to} conflicting results \cite{mo2020audet}.
\del{The PlatEMO manual }\new{A benchmarking software} \cite{tian2017platemo} may use the final population whereas another \cite{biobjective2016brockhoff} tracks all non-dominated points seen during the search process. Here we rely on this latter approach to build sets of non-dominated points for each algorithm.

We \del{maintain an archive $\Arc$ of} \new{record} all points sampled during each algorithm's search process. \new{After aggregating all} sets of non-dominated points from different algorithms we perform a non-dominated sort to obtain the (reference) best set $\Best$, and $f(\Best)$ the so-called Empirical Pareto Front (EPF). This is the set containing all non-dominated points seen so far for the problem at hand as in a real-world context we don't know the true PF.

\subsection{Metrics}

To assess the quality of a set of points $\Points$ against $\Best$ we propose to use metrics computed in the objective space, comparing $f(\Points)$ to $f(\Best)$.

Before each metric is computed, we transform the objective space, each coordinate is scaled from $[\bmin_i, \bmax_i]$ to $[0, 1]$,
computing \del{$\tilde{f}$ as follows:} ${\tilde{f}_i(\x) =(f_i(\x) - \bmin_i)/(\bmax_i - \bmin_i)}$
where $\bmin_i = \min_{\x \in \Best} f_i(\x)$ and $\bmax_i = \max_{ \x \in \Best} f_i(\x)$

\subsubsection{Cardinality}

In our application context we want to know how many points in $\Best$ come from each algorithm. We count the number of non-dominated points obtained by each algorithm. When dealing with a high dimensional objective space this is not a reliable quality indicator, as most points are non-dominated and an algorithm could sample many non-dominated points from only a small connected area of the Pareto Set.

\subsubsection{Hypervolume}

A \new{contemporary} metric is the Hypervolume (HV) \cite{berghammer2012convergence} of the considered set with respect to a reference point $\refpoint$. The maximum HV is achieved by the continuous True Pareto Front. The construction of the best set leads to $\HV_c(\Best) \geq \HV_c(\Points)$ for any set $\Points$ from a single algorithm. The greater the hypervolume, the better the PF is covered. The reference point can be the nadir point (the vector composed with the worst objectives values over the Pareto Front, here $[1,\dots,1]$ after scaling) or another point of interest.
\begin{equation}
\HV_{\refpoint}(\Points) = \text{VOL}(\cup_{\x \in \Points} [\tilde{f}_1(\x), \refpoint_1] \times \dots \times [\tilde{f}_M(\x), \refpoint_M])
\end{equation}
where $\text{VOL(.)}$ is the usual Lebesgue measure. \del{In the PlatEMO framework, w}\new{W}hen the number of objectives $M$ is greater than 4, we do not compute exact hypervolume but rather estimate it by Monte Carlo approximation with 1,000,000 sampled points. We slightly abuse our notation and denote by $\HV_c$ the hypervolume with reference point $\refpoint = c \cdot [1, \, \dots, \, 1]$.

\subsubsection{Generational and Inverted Generational Distances}
GD and IGD \cite{coello2005solving} compare two sets of solutions. Here we always compare a set $\Points$ to the EPF $\Best$.
GD iterates over the set of points to be checked $\Points$:
\begin{equation}
\GD(\Points, \Best) = \frac{1}{\#\Points} \sum_{\x \in \Points} d_{\tf}(\x, \Best)\del{;\; \IGD = \frac{1}{\#\Best} \sum_{\y \in \Best} d_{\tf}(\y, \Points)}
\end{equation}
where ${d_{\tf}(y, \Points) = \min_{\x \in \Points} ||\tf(\x) - \tf(\y)||_2}$ is the distance in the \new{scaled} objective space.
By construction all non-dominated points from $\Points$ also belong to $\Best$, then GD penalizes points that turned out to be dominated when compared to the EPF.

IGD iterates over points on the EPF and checks for closest points in $\Points$. \new{In other words $\IGD(\Points, \Best) = \GD(\Best, \Points)$.}
\del{
\begin{equation}
\IGD(\Points, \Best) = \frac{1}{\#\Best} \sum_{\y \in \Best} d_{\tf}(\y, \Points)
\end{equation}
}
IGD is meaningful when \del{there are enough points in }the reference set $\Best$ \new{provides good coverage of the true Pareto Front}; \del{when the true Pareto Front is known, it makes sense to uniformly sample these points.}
in our context the quality of the EPF can bias the metric.\del{and such requirements can be seen as limits of the performance assessment using the IGD quality indicator.}

\subsection{Algorithms}

 We rely on a previous study \cite{emoo2018li} comparing 13 algorithms on classes of problems with 2, 3 or 7 objectives and select 6 algorithms to run on \Cref{problem_pri}. Parameters setting follows the experimental setup suggested in \cite{emoo2018li}.
 \new{Fine tuning the parameters of each algorithm is beyond the scope of the paper.}

\review{fill algorithms description}

\begin{outline}

	\1 The \emph{Non-dominated Sorting Genetic Algorithm} (NSGA-II) \cite{deb2002nsga} is maybe the most known and used MOEA. It uses Pareto dominance as first sorting procedure and evaluates crowding distance as a second criterion for diversity preservation.
	\1 The \emph{new Multiple Single Objective Pareto Sampling algorithm} MSOPS-II \cite{msopsii2007hughes} was proposed by E. Hughes as a general purpose solver for MaOPs and used for the radar WF problem in \Cref{sec_problem}. MSOPS is not Pareto-based but \new{computes fitness} based on predefined target vectors. MSOPS-II adds automatic target vectors generation and clarified fitness assignment.
	\1 The \emph{Indicator Based EA} (IBEA) \cite{ibea2004zitzler} where the search is guided by the (additive) \epsind{}. This indicator is compliant with the Pareto- dominance relation. The default scaling factor used in the fitness computation is $\kappa = 0.05$. \review{maybe more}
	\1 NSGA-III \cite{deb2014nsgaiii} is a reference-point based evolutionary algorithm following the NSGA-II framework.  Like NSGA-II, it uses first non dominated sorting and then selects points based on a niche-preservation criterion. These niches are meant to bias the search towards the specified reference points. Here we generate $\popsize$ reference points uniformly distributed on the unit hyperplane.
	\1 The \emph{Grid-based EA} (GrEA) \cite{yang2013grid} where the fitness assignement procedure depends on three grid-based criterions. The number of divisions of the objective space in each dimension is $div = 10$.
	\1 \tdea{} \cite{yuan2016theta} where the Pareto dominance is replaced by $\theta$-dominance, here $\theta=5$ is a penalty parameter. Here again we generate $\popsize$ reference points uniformly distributed on the unit hyperplane.
\end{outline}

\section{Experiments}\label{sec_exp}

For all experiments problem dimension is $D = 10$. The initial population \del{is an input to the algorithm, and can be either randomly sampled or given by a domain-specific heuristic. In the following it }is uniformly sampled in $[500, 1500]^D$.
We run each algorithm $\Nruns = 10$ times and each time for a fixed number $\Nfe = 100,000$ of function evaluations. The population size $\popsize$ is also fixed to be the same for each algorithm, i.e. $\popsize = 100$. We aggregate points by algorithm, concatenating data from the different runs, and keeping all non-dominated points to build each set $\Points$. \new{All algorithms are implemented in the PlatEMO library \cite{tian2017platemo}. The software for the MaOP is proprietary and only a Matlab binary file is available. The reported time for one function evaluation is of the order of a millisecond, which allows for a high number of functions calls \cite{radar2007hughes}.}

From our generated data we obtain a first best set of size $217,166$. We also added the set of solutions provided E. Hughes in his original paper \cite{radar2007hughes}. At the end we obtain a final best set  $\Best = 222,667$ out of $229,016$ candidates.
The computational requirements for each algorithm are of the same order, in average between 37.6 and 64.5 seconds in CPU time, except for MSOPS-II which is 4 to 5 times longer.

\subsection{EMO Metrics}

\begin{table*}
	\centering
	\begin{tabular}{|>{\columncolor[HTML]{EFEFEF}}l |l|l|l|l|l|l|l|l|l|}
		\hline
		Algorithm                                & NSGA-II & NSGA-III & IBEA   & GrEA   & MSOPS-II & \tdea &  Hughes \cite{radar2007hughes} & $\Best$ \\ \hline
		$\# \Points$ & 69,512  & 85,887   & 21,344  & 59,699  & 27,700   & 44,798      & 11,850  &  222,667 \\ \hline
		$\# (\Points \cap \Best)$ & 42,406   & \emphb{54,732} & 19,384  & 41,165  & 22,941   & 36,342      &    5,712   &   \\
		( $/ \# \Points$)	 & (61.0\%)  & (63.7 \%)  & \emphb{(90.8 \%)}  & (69.0 \%)  & (82.8 \%)   & (81.1 \%)  &    (48.2 \%)  &  100 \%  \\ \hline
		$\HV_{0.9}(\Points)$                                & 6.94e-4 & 7.83e-4  & \emphb{1.446e-3} & 1.048e-3 & 4.61e-4  & 8.49e-4     & 2.16e-4 & 1.17e-3 \\ 
		($/ \HV_{0.9}(\Best)$) & (41.31\%) & (46.61\%) & \emphb{(86.07 \%)}& (62.38\%) & (27.44\%) & (50.54\%) & (18.46\%) & \\ \hline
		$\HV_1(\Points)$                                & 4.674e-3 & 3.458e-3  & \emphb{7.011e-3} & 5.516e-3 & 4.019e-3  & 3.788e-3    & 2.20e-3 & 7.5e-3 \\
		($/ \HV_{1}(\Best)$) & (62.52\%) & (46.25\%) & \emphb{(93.78\%)} & (73.78\%) & (53.76\%) & (50.67\%) & (29.33\%) & \\ \hline
		$\HV_{1.1}(\Points)$ & 1.3384e-2 & 8.795e-3 & \emphb{1.7343e-2} & 1.4815e-2 & 1.1972e-2 & 9.643e-3
		& 7.40e-3 & 1.92e-2 \\
		($/ \HV_{1.1}(\Best)$) & (69.81\%) & (45.87\%) & \emphb{(90.46\%)} & (77.27\%) & (62.44\%) & (50.29\%) & (38.54\%) & \\ \hline
		$\GD$ & 1.1245e-4 & \emphb{8.7501e-05} & 8.8724e-05 & 1.0369e-4 & 1.1945e-4 &  9.2468e-05 & 3.3103e-04 & 0\\ \hline
		$\IGD$ & \emphb{6.7534e-2} & 1.2664e-1 & 9.1345e-2 & 7.7832e-2 & 9.7094e-2 &  1.4208e-1&  1.389e-1 & 0\\ \hline
	\end{tabular}
	\caption{Summary of all metrics, best performing algorithm for a given metric is in emphasized.\del{ (in order):  Number of non-dominated points obtained during the optimization proces (total of 1,000,000 function evaluations), number of points which stay non-dominated when compared to others in $\Best$, Hypervolume ($\HV$, to be maximized) for three different reference points, Inverted Generational Distance ($\IGD$, to be minimized), Generational Distance ($\GD$, to be minimized). For $\HV$ we also display the ratio in percents of $\HV(\Points) / \HV(\Best)$}}
	\label{table_all}
\end{table*}

Each algorithm dataset consists of $\Nruns \times \Nfe = 1,000,000$ \new{points} from which we compute the different metrics \del{computed for each algorithm}\new{as shown in} \Cref{table_all}.

NSGA-II and NSGA-III are the 2 algorithms that contributed the most to $\Best$ in the number of non-dominated points. Yet they also have the lowest ratios of non-dominated points added to $\Best$ from their own set. This suggests that the NSGA framework does not converge to a fixed set of points on the PF but is rather moving closed to it, finding new non-dominated points in the high dimensional objective space.
This cyclic behaviour is well known in the EMO litterature \cite{berghammer2012convergence}.
For comparison, IBEA gives a smaller set $\Points$ of non-dominated points, but a large part of it contributes to $\Best$ (90.8\%).
Note that, not only we improved the original dataset by increasing its size, but half of the points provided by the original experiment \cite{radar2007hughes} are now dominated by $\Best$.

Hypervolume is computed for 3 different reference points $c \in \{0.9, 1, 1.1\}$ and approximation variance has been checked to be acceptable for comparison.
We always give the hypervolume of the set $\Best$ as a reference, and percentages comparing the hypervolume of any set $\Points$ against $\Best$.
The idea to look at many reference points is the following: it is known that the hypervolume quality indicator, when consider high dimensional spaces, tends to promote set of points that are mainly located on the boundary of the Pareto Front (as they have a greater contribution to the integral). By comparing the obtained measures (normalized by $\HV_c(\Best)$), we may have a guess about which algorithm is able to generate solution sets being well spread on the PF.
As an example, the hypervolume decrease from 62.52\% to 41.31\% for NSGA-II when switching the ref. point parameter $c$ from 1 to 0.9.
The same remark holds for MSOPS-II (53.76\% to 27.44\%).
The hypervolume of IBEA is much more stable in all situations, and is also the best performing for any hypervolume measure, whereas NSGA-III is for IGD and NSGA-II for GD. 

Therefore no algorithm overperforms all others for all metrics considered here.

\subsection{Problem insights}

We propose to visualize the set $\Best$ to provide a better understanding of the trade-offs involved in radar WF design. \new{From a practical point of view, we are only interested in the \emph{realistic} solutions for which all \emph{original} objectives are positive \cite{radar2007hughes}. The following analysis is done only over the set $\Best_{+} = \{\x \in \Best;\,f_k(\x) > 0,\,k=1,\,\dots,\,M\}$. We have $\#\Best_{+} = 111,289$.}
\del{Recall that, for objectives 1 to 8, values are negative because we require minimization of the MaOP; and objective 9 (dwell time) was added an offset of 50 ms.}

\begin{figure*}
	\includegraphics[width=\textwidth, height=.20\textheight, trim= 30 30 30 30]{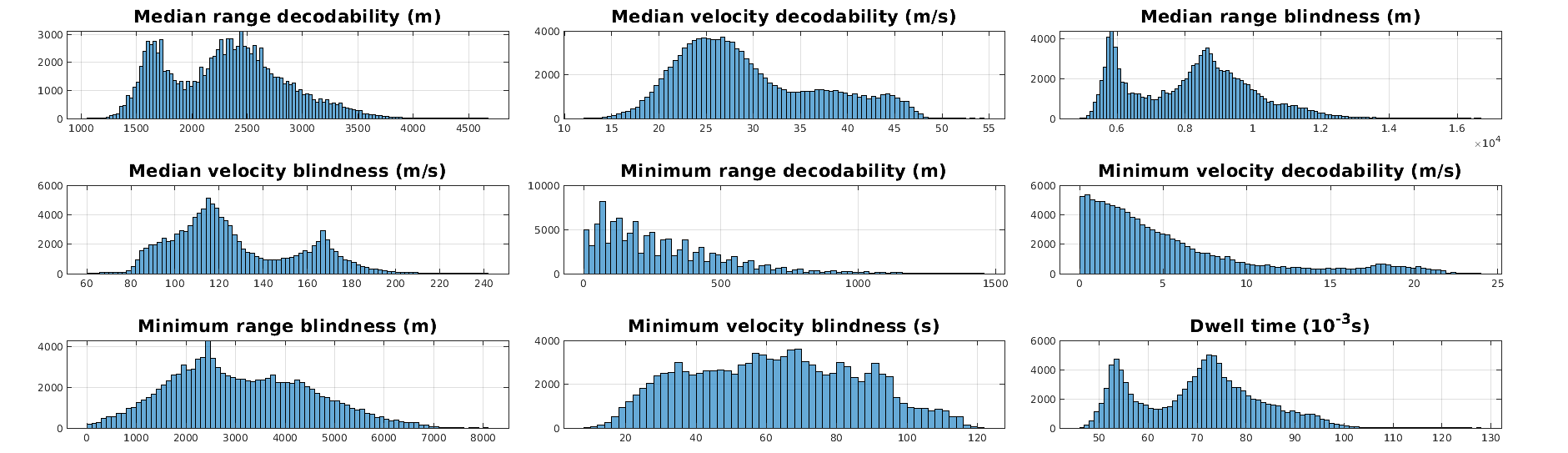}
	\caption{Histogram of objective values in $\Best_{+}$, $D=10$, decodability and blindness to be maximized, dwell time to be minimized \done{figure globalement plus petite, valeur globalement plus grosses}}
	\label{histo_objectives}
\end{figure*}

The distribution of each objective's values in \Cref{histo_objectives}. A remark is the multimodality of distribution for several objectives. But it is not clear if these modes come from exploitation of specific regions of the search space due to algorithms bias or particular shape of the Pareto Front.
More sophisticated methods exist to better restore information from a high dimensional Pareto Front with different guarantees about what is preserved from the original space \cite{visualization2019filipic}. 
\review{plus ici}

\del{If we look at the PF in 2-dimensional subspaces, it is clear that}\new{Investigating 2-dimensional cuts of the EPF, we notice that the} two objectives median velocity blindness ($f_4$) and dwell time ($f_9$) are non-conflicting. This was also detected in a study about Objective Reduction in Multi-objective optimization \cite{brockhoff2009reduction}.
Given the radar system is frequency-hopping, the $n$-th ambiguous velocity is $V_a(n) = \lambda_n/(2x_n)$ where ${\lambda_n = c /(10^{10} - n.3.10^7)}$ is the wavelength and $x_n$ the PRI of the $n$-th pulse, and $c$ the speed of light.
Small dwell time implies small PRIs around 50 \musec. \del{This results in}\new{The consequences are} high ambiguous velocities (250 to 300 $\text{ m.s}^{-1}$). On the contrary, high PRIs (closer to 150 \musec) mean smaller ambiguous velocities (around 100 to 120 $\text{ m.s}^{-1}$).
Since targets are searched in the velocity range $\pm 1500 \text{ m.s}^{-1}$ this explains the observed difference by a factor 2 to 3 (omitting other second-order effects).

\subsection{Comparison with existing methods}

\begin{figure}
	\includegraphics[width=\linewidth, trim=20 20 20 20]{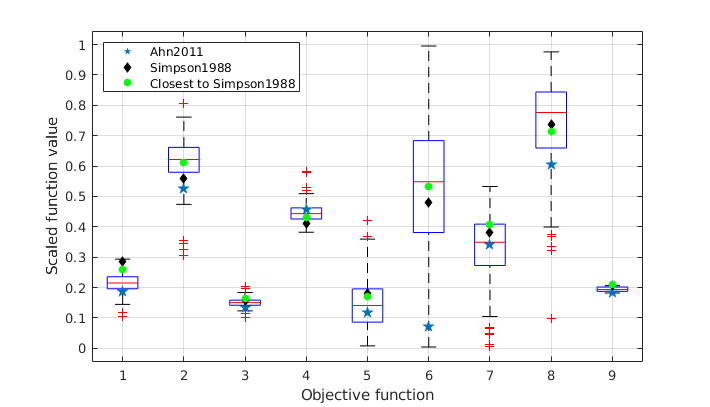}
	\caption{Boxplots of objective values in $\Points$(IBEA), $D=8$, solutions are filtered such that the original objective values are all positive, then scaled to $[0,1]$  \del{following \cref{eq_scaling}} hence we seek to \emph{maximize} $f_1$ to $f_8$ and \emph{minimize} $f_9$.
	We show only the solutions leading to a dwell time between 45.4 and 46.5 ms.
	Previously found solutions $\x_s$ (in black) and $\x_a$ (in blue) are added for comparison \cite{ahn2011medium, simpson1988prf}.
	To show that the obtained set provides solutions which are closed to the existing ones, we display $\x_c$ (in green) the closest point to $\x_s$ from $\Points$(IBEA) where $d_{\tf}(\x_c, \Points(\text{IBEA})) = 0.0887$. It has a dwell time of 46.6 ms.}	\label{boxplot_smalldwell}
\end{figure}

We compare two PRI vectors from the literature with our findings with respect to the baseline problem considered here. Simpson \cite{simpson1988prf} gives a vector of 8 PRFs that we convert to PRIs:
\del{$$\x_s &= [19.61, 17.54, 15.87, 15.15, 14.49, 12.82, 11.11, 10.42]\tran \text{ kHz}$$}
\begin{equation}
\x_s = [510, 570, 630, 660, 690, 780, 900, 960]\tran (\times 0.1\mu\text{s})
\end{equation}
Ahn et al. give a vector of 8 PRIs with a time quantization of 1 $\mu$s:
\begin{equation}
\x_a = [510, 530, 590, 620, 690, 720, 910, 940]\tran (\times 0.1\mu\text{s})
\end{equation}
In this comparison we use only IBEA as the most promising algorithm and repeat the previously defined experimental setup to generate $\Nruns \times \Nfe = 1,000,000$ points with $D=8$.

We obtain a set of 24,259 non-dominated points. Among these, 17,005 points are realistic. We use this latter set for scaling.
In order to compare the solution set with $\x_s$ and $\x_a$ we display only the points in the dwell time interval between $\x_s$ and $\x_a$ in \Cref{boxplot_smalldwell}.
Note that the PRIs proposed by Simpson was found to be non-dominated with respect to $\Points$(IBEA), whereas the PRIs proposed by Ahn et al. was dominated.
This may be due to the lack of precision resulting from greater time quantization.

\new{Interestingly, these solutions were found with different parameter settings for the radar system, for example the duty cycle which is exactly 1.5\% in \cite{ahn2011medium} and averaged to 1.4\%  in \cite{simpson1988prf}. Despite this, they operate at medium PRF and share common goals. We find that $\vec{\x_s}$ is still Pareto-optimal for our problem \cite{radar2007hughes}, suggesting that these problems are similar.}

Simpson's solution $\x_s$ is also better performing than $\x_a$ for most of the objectives considered here, except $f_4$ (median velocity blindness), with a slightly greater dwell time (46.5 ms for $\x_s$ vs 45.4 ms for $\x_a$). 
IBEA found 321 points in this dwell time interval.
The solution set obtained with IBEA show possible improvements for all objectives.\todo{un com quantitatif ici ?}
From such set the radar system can dynamically select PRFs better suited to the operational scenarii.

\done{revoir ça avec les objectifs réels, notamment le dwell time} 
\done{Pourquoi le dwell time est de 50ms quand les 8 PRIs sont autour de 100\musec{}.}

\section{Summary and conclusion}\label{sec_future}

\done{radar plutot que radar}
\done{Waveform (WF) pour gagner de la place}

We considered several efficient evolutionary algorithms to solve Many-objective optimization problems and benchmarked them on the radar WF design problem with 10 PRIs.
We obtained a large collection of non-dominated points which describes the unavoidable trade-offs when designing such radar signals.
We also compared the obtained WFs with 8 PRIs to existing single solutions from the literature where our solution set appears more diverse.
This work could be extended by searching  for other numbers of PRIs or refining the objectives.
This is an \emph{offline} search procedure: the solvers are not aimed at running on the radar hardware, but the obtained PRIs database could be used with some specific rules to switch the radar mode while operating. \new{The radar system can also sequentially select PRIs with a diversity criterion to make adverse detection complicated.}
We also illustrated the difficulty of performance assessment when many objectives are involved.
\new{Real-world concerns suggest to integrate user preferences and constraints: since a radar WF is useful only if all objectives have positive values, the interesting MaOP is constrained. Handling such constraints on the objective functions can be done through weighted hypervolume or modified \epsind{} to guide the optimization process and assess performance.}

\bibliographystyle{IEEEtran}
\bibliography{references} 

\begin{thebibliography}{10}
\providecommand{\url}[1]{#1}
\csname url@samestyle\endcsname
\providecommand{\newblock}{\relax}
\providecommand{\bibinfo}[2]{#2}
\providecommand{\BIBentrySTDinterwordspacing}{\spaceskip=0pt\relax}
\providecommand{\BIBentryALTinterwordstretchfactor}{4}
\providecommand{\BIBentryALTinterwordspacing}{\spaceskip=\fontdimen2\font plus
\BIBentryALTinterwordstretchfactor\fontdimen3\font minus
  \fontdimen4\font\relax}
\providecommand{\BIBforeignlanguage}[2]{{%
\expandafter\ifx\csname l@#1\endcsname\relax
\typeout{** WARNING: IEEEtran.bst: No hyphenation pattern has been}%
\typeout{** loaded for the language `#1'. Using the pattern for}%
\typeout{** the default language instead.}%
\else
\language=\csname l@#1\endcsname
\fi
#2}}
\providecommand{\BIBdecl}{\relax}
\BIBdecl

\bibitem{simpson1988prf}
J.~{Simpson}, ``Prf set selection for pulse-doppler radars,'' in \emph{IEEE
  Region 5 Conference, 1988: 'Spanning the Peaks of Electrotechnology'}, 1988,
  pp. 38--44.

\bibitem{alabaster2003medium}
C.~Alabaster, E.~Hughes, and J.~Matthew, ``Medium prf radar prf selection using
  evolutionary algorithms,'' \emph{Aerospace and Electronic Systems, IEEE
  Transactions on}, vol.~39, pp. 990 -- 1001, 08 2003.

\bibitem{ahn2011medium}
S.~Ahn, H.~Lee, and B.~Jung, ``Medium prf set selection for pulsed doppler
  radars using simulated annealing,'' \emph{IEEE National Radar Conference -
  Proceedings}, 05 2011.

\bibitem{radar2007hughes}
E.~Hughes, ``{Radar Waveform Optimisation as a Many-Objective Application
  Benchmark},'' in \emph{Evolutionary Multi-Criterion Optimization},
  S.~Obayashi, K.~Deb, C.~Poloni, T.~Hiroyasu, and T.~Murata, Eds.\hskip 1em
  plus 0.5em minus 0.4em\relax Berlin, Heidelberg: Springer Berlin Heidelberg,
  2007, pp. 700--714.

\bibitem{kinghorn1997decodability}
A.~Kinghorn and N.~Williams, ``The decodability of multiple-prf radar
  waveforms,'' \emph{IET Conference Proceedings}, pp. 544--547(3), January
  1997.

\bibitem{mo2020audet}
C.~Audet \emph{et~al.}, ``{Performance indicators in multiobjective
  optimization},'' Feb. 2020, working paper or preprint.

\bibitem{tian2017platemo}
Y.~Tian, R.~Cheng, X.~Zhang, and Y.~Jin, ``{PlatEMO}: A {MATLAB} platform for
  evolutionary multi-objective optimization,'' \emph{IEEE Computational
  Intelligence Magazine}, vol.~12, no.~4, pp. 73--87, 2017.

\bibitem{biobjective2016brockhoff}
D.~B. et~al., ``Biobjective performance assessment with the {COCO} platform,''
  \emph{CoRR}, vol. abs/1605.01746, 2016.

\bibitem{berghammer2012convergence}
R.~B. et~al., ``Convergence of set-based multi-objective optimization,
  indicators and deteriorative cycles,'' \emph{Theoretical Computer Science},
  vol. 456, pp. 2 -- 17, 2012.

\bibitem{coello2005solving}
C.~Coello \emph{et~al.}, ``{Solving Multiobjective Optimization Problems Using
  an Artificial Immune System},'' pp. 163--190, 2005.

\bibitem{emoo2018li}
K.~{Li} \emph{et~al.}, ``{Evolutionary Many-Objective Optimization: A
  Comparative Study of the State-of-the-Art},'' \emph{IEEE Access}, vol.~6, pp.
  26\,194--26\,214, 2018.

\bibitem{deb2002nsga}
K.~{Deb}, A.~{Pratap}, S.~{Agarwal}, and T.~{Meyarivan}, ``A fast and elitist
  multiobjective genetic algorithm: Nsga-ii,'' \emph{IEEE Transactions on
  Evolutionary Computation}, vol.~6, no.~2, pp. 182--197, 2002.

\bibitem{msopsii2007hughes}
E.~Hughes, ``{MSOPS-II: A general-purpose Many-Objective optimiser},'' in
  \emph{2007 IEEE Congress on Evolutionary Computation}, 10 2007, pp. 3944 --
  3951.

\bibitem{ibea2004zitzler}
E.~Zitzler \emph{et~al.}, ``{Indicator-Based Selection in Multiobjective
  Search},'' in \emph{Conference on Parallel Problem Solving from Nature (PPSN
  VIII)}, 09 2004, pp. 832--842.

\bibitem{deb2014nsgaiii}
K.~{Deb} \emph{et~al.}, ``{An Evolutionary Many-Objective Optimization
  Algorithm Using Reference-Point-Based Nondominated Sorting Approach Part I:
  Solving Problems With Box Constraints},'' \emph{IEEE Transactions on
  Evolutionary Computation}, vol.~18, no.~4, pp. 577--601, 2014.

\bibitem{yang2013grid}
S.~Y. et~al., ``A grid-based evolutionary algorithm for many-objective
  optimization,'' \emph{IEEE Transactions on Evolutionary Computation},
  vol.~17, no.~5, pp. 721--736, 2013.

\bibitem{yuan2016theta}
Y.~Y. et~al., ``A new dominance relation-based evolutionary algorithm for
  many-objectiveoptimization,'' \emph{IEEE Transactions on Evolutionary
  Computation}, vol.~20, no.~1, pp. 16--37, 2016.

\bibitem{visualization2019filipic}
B.~Filipi\v{c} \emph{et~al.}, ``{Visualization in Multiobjective
  Optimization},'' in \emph{Proceedings of the Genetic and Evolutionary
  Computation Conference Companion}, ser. GECCO '19.\hskip 1em plus 0.5em minus
  0.4em\relax New York, NY, USA: Association for Computing Machinery, 2019, p.
  951–974.

\bibitem{brockhoff2009reduction}
D.~Brockhoff and E.~Zitzler, ``Objective reduction in evolutionary
  multiobjective optimization: Theory and applications,'' \emph{Evolutionary
  Computation}, vol.~17, no.~2, pp. 135--166, 2009, pMID: 19413486.

\end{thebibliography}

\end{document}